# A Comparison of Veterans with Problematic Opioid Use Identified through Natural Language Processing of Clinical Notes versus Using Diagnostic Codes

T. Elizabeth Workman, PhD,[1,2,*] Joel Kupersmith, MD,[3] Phillip Ma, MD,[1,2] Christopher Spevak, MD, MPH, JD,[3] Friedhelm Sandbrink, MD,[1] Yan Cheng, PhD,[1,2] Qing Zeng-Treitler, PhD[1,2]

[1]Washington DC VA Medical Center, Washington, DC, USA
[2]Biomedical Informatics Center, the George Washington University, Washington, DC, USA
[3]Georgetown University School of Medicine, Washington, DC, USA
*corresponding author (lizworkman@gwu.edu)

## Abstract

**Background:** Electronic health records (EHRs) are a data source for opioid research. Opioid use disorder is known to be under-coded as a diagnosis, yet problematic opioid use can be documented in clinical notes.

**Objectives:** Our goals were 1) to identify problematic opioid use from a full range of clinical notes; and 2) to compare the characteristics of patients identified as having problematic opioid use, exclusively documented in clinical notes, to those having documented ICD opioid use disorder diagnostic codes.

**Materials and Methods:** We developed and applied a natural language processing (NLP) tool to the clinical notes of a patient cohort (n=222,371) from two Veteran Affairs service regions to identify patients with problematic opioid use. We also used a set of ICD diagnostic codes to identify patients with opioid use disorder from the same cohort. We compared the demographic and clinical characteristics of patients identified only through NLP, to those of patients identified through ICD codes.

**Results:** NLP exclusively identified 57,331 patients; 6,997 patients had positive ICD code identifications. Patients exclusively identified through NLP were more likely to be women. Those identified through ICD codes were more likely to be male, younger, have concurrent benzodiazepine prescriptions, more comorbidities, more care encounters, and less likely to be married. Patients in the NLP and ICD groups had substantially elevated comorbidity levels compared to patients not documented as experiencing problematic opioid use.

**Conclusions:** NLP is a feasible approach for identifying problematic opioid use not otherwise recorded by ICD codes. Clinicians may be reluctant to code for opioid use disorder. It is therefore incumbent on the healthcare team to search for documentation of opioid concerns within clinical notes.

## Introduction

Opioid misuse is a serious, escalating public health issue. Since 1999, there have been 932,000 overdose deaths in the United States, mainly driven by opioids [1]. This has been exacerbated by the COVID pandemic; in 2020 preventable opioid overdose deaths increased 41% [2]. Veterans of the U.S. military may especially be harmed by the misuse of opioids. Veterans are more likely to experience opioid poisoning mortality [3], and face increasing rates of opioid use disorder (OUD) [4] and opioid overdose death [5].

To prevent overdose and other opioid complications, it is imperative to identify patients experiencing problematic opioid use. Electronic health records (EHRs) can serve as a data source, and include structured and unstructured data, the latter in the form of free text clinical notes. Relying only on structured data is insufficient because opioid overdose is under-coded [6], and structured data are insufficient for identifying OUD [7]. Identifying patients experiencing problematic opioid use through EHR data could provide early detection of cases that are otherwise currently missed.

Natural language processing (NLP) can extract information from free text and has been applied to clinical notes to identify documentation of opioid use issues, usually for specific clinical settings or patient types. NLP systems generally consist of hard-coded rules, trained machine learning models, or both. Blackley et al. randomly selected emergency department notes, inpatient progress notes, and previous hospital discharge summaries to develop a rule-based model plus several machine learning models to classify inpatient encounters by OUD status, with best performance achieved by a random forest classifier (97% accuracy) [8]. Using the EHR data and notes of chronic opioid therapy patients, Carrell et al. developed an NLP application to identify OUD patients not otherwise identified through ICD codes, with a focus on maximizing sensitivity, yielding a 34.6% false positive rate for documents, and a 41% false positive rate for patients [9]. Afshar et al. used inpatient notes for patients screened for illicit drug use to develop a convolutional neural network classifier, achieving 81% sensitivity and 72% positive predictive value (PPV) [10]. Zhu et al. developed a rule-based system achieving 98.5% precision, 100% recall, 99.2% F1, but also limited their study to the notes of chronic opioid therapy users [11]. In research pursuant to data de-identification, Sharma et al. [12] developed diverse classifiers using inpatient encounters, experimenting with several different types of features. For models having n-gram features, a convolutional neural network achieved an F1 score of 84%, 94% PPV, 75% recall, 98% specificity, and 88% negative predictive value. An n-gram is a word sequence in text, where the n references the word count. For example, "prescription" is a unigram, and "primary provider" is a bigram.

Many of these prior studies targeted patients who already had a recognizable risk for problematic opioid use, which does not take full advantage of information for all patients and all settings available in EHR data sources. This begs the question of whether the characteristics of patients who received relevant diagnostic codes are different from all patients, from all settings, identified as at risk for problematic opioid use using clinical notes. To answer this, using notes from all encounter types, and all types of patients would provide an expanded view of problematic opioid use documentation in EHR data.

The objectives of this study were to identify problematic opioid use in notes from all available clinical settings and patients, and analyze the differences between patients identified exclusively through NLP classification, and patients identified using ICD codes, concerning demographics, selected comorbidities, prescription data, and healthcare encounters. Our initial hypothesis was that there would be a difference between these two groups. To pursue this we developed an open-source NLP tool (https://github.com/GWU-BMI/opioids-nlp) to identify problematic opioid use in clinical notes from a broad range of clinical settings and patients. We also analyzed structured data to identify patients receiving an OUD ICD code. Patient EHR records from the U.S. Department of Veterans Affairs (VA), stored within the VA central data warehouse (CDW), served as the data source. We accessed these data for two VA service regions within the secure VINCI research platform [13].

## Materials and Methods

*Study Cohort*

We identified a cohort of 222,371 patients with at least 2 outpatient encounters within the Baltimore, Maryland VA station or the Washington, DC VA station between 1 Jan 2012 and 31 Dec 2019. A VA station is a regional service area that typically includes at least one hospital, and additional resources such as outpatient clinics. These patients had a total of 81,129,781 notes from all clinical settings over that period.

*Annotation Guideline*

To better understand problematic opioid use documentation in clinical notes, two members of the research team independently annotated 196 snippets containing 46 relevant keywords or key phrases (collectively noted as "key phrases"), including generic and trade names for opioid drugs, and phrases like "tapered", "withdrawal", and "opioid abuse" for positive problematic opioid use, suspected problematic opioid use, and negative documentation. This initial list was assembled using findings from previous work and professional clinical knowledge [14, 15]. A positive classification arose from documentation suggesting current abuse, overuse, or addiction to prescribed or illicit opioids. Negative documentation was where a key phrase was present, but there was no indication of problematic opioid use in the text. Each snippet consisted of the key phrase, and the 50 words before it and after it in the document. The snippets were extracted from randomly selected notes containing the key phrases. After calculating inter-rater agreement (86.73%), the two team members reviewed and discussed the results. They then independently reviewed 185 new snippets, measured inter-rater agreement (97.31%), and again discussed the results. They then identified the most relevant key phrases for problematic opioid use documentation.

*NLP Tool Development*

The research team developed an NLP tool that combines supervised machine learning and rule-based pattern recognition to classify snippets. The machine learning component utilizes the position of the key phrase in the text (in cases where there are multiple words in the key phrase, the position of the first word in the key phrase is used), and n-grams as features. The text of all snippets serving as input to the machine learning component was preprocessed by transforming it to lower case, then it was tokenized, and non-alphanumeric characters were removed. Using 582 annotated snippets classified by the research team, a support vector machine (SVM) trained using as features 946 unigrams that consisted entirely of letters, and 474 bigrams where the first word consisted of letters, and the second word consisted of letters, numbers, or a combination of letters and numbers, yielded the best initial performance out of several different algorithms tried, (SVM, Random Forest, AdaBoost). All experiments utilized the python scikit-learn package, version 1.0.2. To assess performance, we measured precision/PPV and sensitivity/recall, using an 80%/20% train/test split of the data. Splitting the data in this fashion enables testing using "unseen" data, i.e. data not used to train a given model. The best SVM achieved 94% precision/PPV and 84% sensitivity/recall, whereas the best random forest model achieved 89% precision/PPV and 78% recall/sensitivity, and the best Adaboost achieved 85% precision/PPV and 80% sensitivity/recall. Snippets annotated as having suspected problematic opioid use were aggregated with snippets annotated as having positive problematic opioid use to serve as positive training examples; snippets annotated as negative documentation of problematic opioid use were used as negative training examples.

To automatically identify template data and standard language found in opioid documentation, we also developed a library of 145 regular expressions. A regular expression is a group of specialized characters designed to match specific types of text. For example, a regular expression like '\[\s*x\s*\]' could be used to match a questionnaire element like "[x]" and account for potential white space around "x". Examples of relevant note content that can be identified using regular expressions includes statements like "former opioid dependence", "sister abuses hydrocodone", and template data like "[x] substance abuse and/or dependence". In these examples, the phrase "former opioid dependence" indicates opioids are a former, i.e. not a current issue for the patient, and "sister abuses hydrocodone" references someone other than the patient. The phrase "[x] substance abuse and/or dependence" is from a questionnaire that directly pertains to the patient at the time it was recorded. To fine-tune the library of expressions, 160 additional documents containing the final key phrases – 40 each from the even years of the study – were randomly selected and reviewed for additional relevant patterns.

Rule-based pattern recognition and machine learning classification provided the operative power of the NLP tool. It received snippets containing the key phrases as input. It first analyzed their content for patterns recognized by the library of regular expressions in the rule-based classifier. This module used a sequential voting system to classify snippets containing these patterns. Regular expressions were grouped by what they identified: absolute positive patterns, canceling patterns, general positive patterns, and neutral patterns. First, snippets containing any absolute positive expressions like "patient continues to struggle with opioid abuse" were classified as positive and transferred to output. Second, the remaining snippets containing any canceling expressions like "former opioid dependence" (i.e., not current) were then classified as negative and transferred to output. Third, the remaining snippets containing any general positive expressions like "opioid abuse" were then classified as positive and transferred to output. Finally, the remaining snippets containing neutral patterns like "sister abuses hydrocodone" and "cannot drive or operate machinery" (general instruction content often accompanying opioid prescriptions) were then classified as negative and transferred to output. Figure 1 illustrates this process.

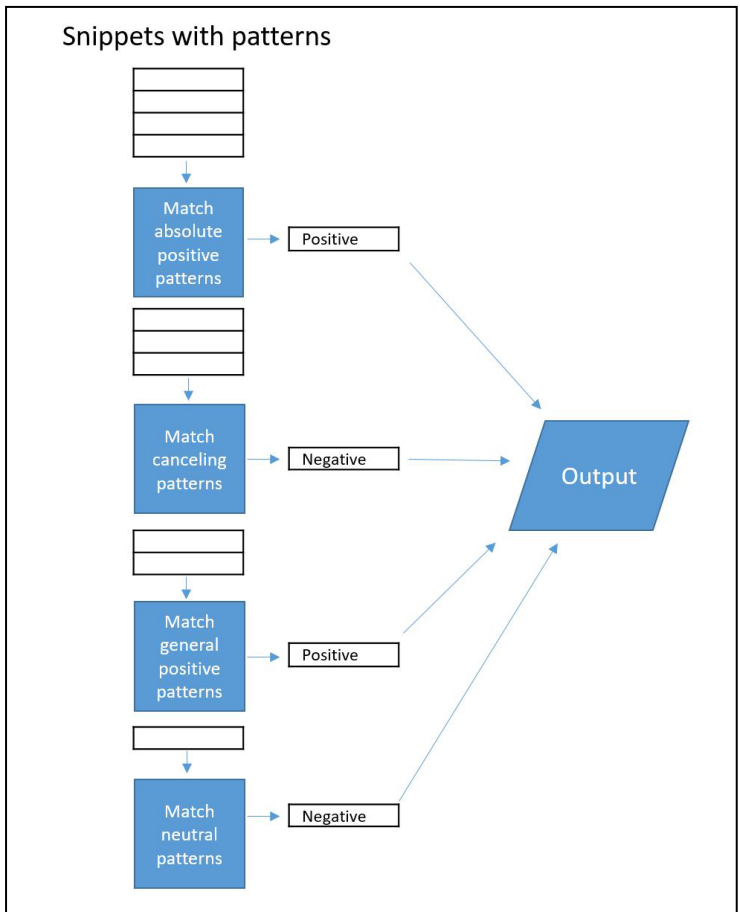

Figure 1. Rule-based classification. Beginning at the top, snippets with relevant patterns are classified sequentially according to pattern type.

After the rule-based classifier had classified all snippets with relevant patterns, the trained SVM model classified the remaining snippets after the preprocessing described earlier. Figures 2 illustrates the NLP tool's overall sequential procedure of classifying snippets.

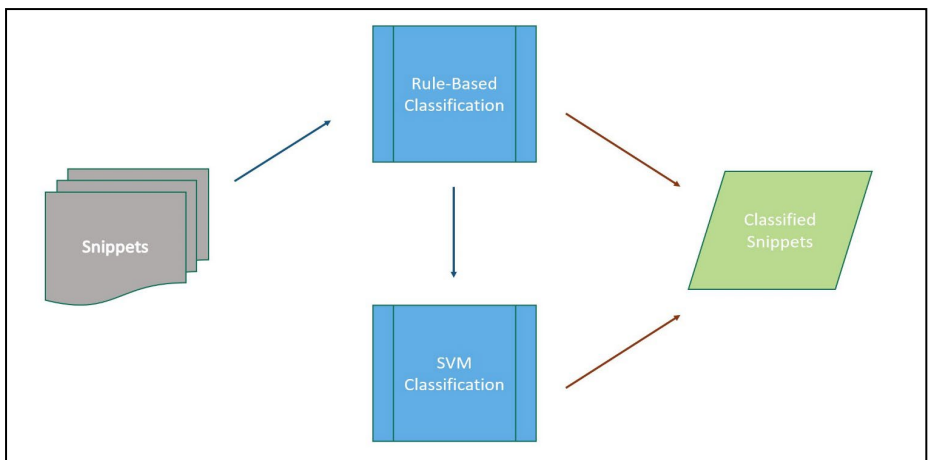

Figure 2. NLP tool operations. After the rule-based classifier has classified all snippets with relevant patterns, the SVM classifies the remaining snippets.

*Evaluation*

A different, unseen test set of 161 annotated snippets not used in the NLP tool's development was used to evaluate the classifier. These snippets consisted of randomly selected data containing one of the key phrases, including randomly selected data containing template markup from the odd years of the study. None of the 161 snippets were used to train the model. By including template data from odd years in testing, we also included potentially new template patterns for the testing process. We measured recall/sensitivity, specificity, precision/PPV, and overall accuracy.

*Classifying Cohort Notes*

We established a minimum performance threshold of 85% for each of the four metrics, which has been recognized as a useful performance threshold in machine learning [16]. Once this was achieved, we applied the classifier to all cohort clinical notes that contained the relevant key phrases.

*Statistical Analysis*

For each of the patient groups we retrieved demographic data, selected comorbidity and prescription data, and outpatient visit counts. The main outcome was problematic opioid use, and how it had been identified. We conducted a two-way t test to compare average ages at index date and average outpatient encounter visits after the index date, and Chi-square tests for the baseline comorbidities, prescription data, and the rest of the demographic data. The index date for each patient was the first date of problematic opioid use documentation, either through ICD code or the NLP tool. These analyses were carried out using SAS, version 9.4 (Cary, NC). Because of the large sample size (N = 222,371), a small difference between groups may be statistically significant. Therefore, we also calculated the absolute standardized difference (ASD) for all variables. An ASD > 10% indicated imbalanced characteristics between two groups [17, 18].

*Grouping Patients by Identification Method*

We used the classifier output to identify patients having problematic opioid use as indicated by NLP. We identified patients diagnosed with OUD as indicated by ICD codes (ICD-CM-9: 304.00, 304.70. 305.50; ICD-CM-10: F11) using the results of the structured data analysis. Patients identified as having problematic opioid use only by the classifier (referred to as *NLP Only*) constituted one group for analysis. Patients identified as having problematic opioid use as indicated by ICD code (referred to as *All ICD*) constituted the other group for analysis. Therefore, by using this strategy, some of the All ICD group patients may also have had one or more snippets identified as positive by the classifier, but none of the NLP Only patients received a relevant ICD code in the study period, and both groups were mutually exclusive. For comparative purposes, we also identified cohort members not identified as having problematic opioid use by either ICD code or NLP (referred to as *No Problematic Opioid Use*).

*Prominent Note Types among Patient Groups*

As an additional measure to shed light on differences between patients identified as having problematic opioid use only through NLP (NLP Only), and those who were identified through relevant ICD codes that also had clinical notes positive for problematic opioid use (referred to as *NLP/ICD*), all clinical notes identified by NLP as positive for problematic opioid use for patients in both groups were retrieved. We analyzed multiple properties of the notes and their snippets.

## Results

*Key Phrases*

In building the annotation guideline, 36 key phrases relevant to problematic opioid use were identified. Table 1 contains these key phrases.

| Key phrases | | | | | |
|---|---|---|---|---|---|
| abstral | duragesic | hysingla | methadose | oxaydo | withdrawal |
| actiq | exalgo | kadian | morphine | oxycodone | zohydro |
| demerol | fentanyl | lorcet | norco | oxycontin | opioid dependence |
| dependence | fentora | lortab | opiate | percocet | polysubstance abuse |
| dilaudid | hydrocodone | meperidine | opiate abuse | roxicet | substance abuse |
| dolophine | hydromorphone | methadone | opioid | vicodin | substance dependence |

Table 1. Key phrases

*Classifier Performance*

The classifier achieved 96.6% specificity, 90.4% precision/PPV, 88.4% sensitivity/recall, and 94.4% accuracy. This performance satisfied the threshold we had previously established. We also felt it was sufficiently comparable to that of other studies, considering the expanded task of accommodating data from all clinical settings and patients.

*Clinical Note Classification*

The NLP tool was used to classify 3,521,637 notes of cohort patients (notes containing one or more of the key phrases). These notes contained 8,804,031 snippets, each including one of the key phrases. Table 2 contains these results on a document-basis. Table A in the Appendix includes counts for each key phrase in the snippets. Table B in the Appendix includes the number of patients having an occurrence of each key phrase in a snippet.

| Element | Total |
|---|---|
| Years (2012 - 2019) | 8 |
| Key phrases | 36 |
| Total notes | 3,521,637 |
| Total snippets | 8,804,031 |
| positive snippets | 1,885,642 |
| negative snippets | 6,918,389 |
| Mean snippets per document | 2.9 |

Table 2. Clinical note classification, document-basis.

Table 3 contains example text spans from positive and negative snippets and the classification method used. Key phrases are highlighted in yellow. The positive examples are indicative of the types of positive documentation sought according to the study's goals.

| Positive for Problematic Opioid Use and Classification Method | | Negative for Problematic Opioid Use and Classification Method | |
|---|---|---|---|
| …substance abuse treatment…heroin last used: "yesterday"… | Machine learning | …pt has pain mostly at night was on lorcet and tried to change to morphine but since she developed rash… | Machine learning |
| …4. low back pain…5. opioid dependence…6. homeless single person… | Regular expression | …hydromorphone 4mg tab take one tablet every four active hours when needed for pain… | Regular expression |
| …opioid dependence (icd-9-cm 304.00)… | Regular expression | …family hx of substance abuse… | Regular expression |
| Alludes to the possibility of self medicating on the street…opiate withdrawal | Machine learning | …patient requested no lortab… | Machine learning |
| …would not receive prescription for morphine and oxycodone until next month…reiterated multiple times that taking additional doses of opiates was a patient safety issue and would not be tolerated… | Machine learning | …continue tylenol and oxycodone as needed per home regimen… | Machine learning |
| ...allergies: darvon, periactin, phenothiazine/related antipsychotics, demerol…opioid dependence (icd-9-cm 304.00) | Regular expression | …9) hydromorphone inj, soln active…give: 0.5 mg/0.5ml ivp q2h prn…for pain… | Regular expression |

Table 3. Positive and negative snippet text examples, with classification method

The positive examples identified through regular expressions include "opioid dependence" in an enumerated problem list, an "opioid dependence" included as a current diagnosis, plus a snippet that included both opioid dependence in the patient's current problem list, and an allergy to Demerol. In this last example, the "opioid dependence (icd-9-304.00)" was identified as an "absolute positive expression" first, so the snippet was classified as positive, despite also documenting an allergy to an opioid drug, which constituted a "neutral expression". The positive examples identified through the SVM include "substance abuse", "oxycodone" (and "morphine"), and "withdrawal". The three negative examples identified through regular expressions include directions for taking medication (lortab), a family history of substance abuse (that does not pertain directly to the patient) and an opioid medication (hydromorphone) included in a list of current medications. The negative machine learning examples include a patient's experience of taking the drug where she developed a rash, a patient that did not request the drug, and directions for taking medications (including oxycodone).

*Problematic Opioid Use in Patients*

The NLP tool identified 63,574 patients that received 1 or more positive snippets for problematic opioid use. Of all patients positively classified through NLP, 57,331 received a positive classification exclusively by that method (NLP Only). Within the cohort, 6,997 patients had received 1 or more ICD diagnostic codes for OUD (All ICD) (6,243 patients in the All ICD group also had one or more positive snippets). There were differences between the two groups regarding demographic attributes (Table 4), comorbidities, prescription data, and outpatient visits (Table 5). In these tables, ASD > 10% (bold text) indicates significant imbalanced characteristics between two groups. For comparison, the values of cohort members not classified by NLP or ICD code (No Problematic Opioid Use) are included, along with p-values and ASD as compared to the NLP Only group in Tables 4 and 5.

|  | All ICD | NLP Only | p-value (All ICD vs NLP Only) | ASD (All ICD vs NLP Only) | No Problematic Opioid Use | p-value (NLP Only vs No Problematic Opioid Use) | ASD (NLP Only vs No Problematic Opioid Use) |
|---|---|---|---|---|---|---|---|
| N | 6997 | 57331 |  |  | 158,043 |  |  |
| **Gender %** |  |  | <0.0001 |  |  | <0.0001 |  |
| M | 93% | 82% |  | **34** | 84.9% |  | 8 |
| F | 7% | 18% |  | **34** | 15.1% |  | 8 |
| **Mean Age/Standard deviation** (at year patient entered cohort) | 53.3/12.2 | 55.4/16.1 | <0.0001 | **15** | 58.8/18.7 | <0.0001 | **17** |
| **Marital Status %** |  |  | <0.0001 |  |  | <0.0001 |  |
| Married | 25.7% | 38.5% |  | **28** | 50.2% |  | **24** |
| Divorced | 31.6% | 25.8% |  | **13** | 17.1% |  | **21** |
| Never Married/Single | 26.5% | 22.8% |  | 9 | 15.6% |  | **18** |
| Widowed | 4.5% | 5.1% |  | 3 | 6.9% |  | 8 |
| Separated | 11.3% | 6.5% |  | **17** | 3.2% |  | **16** |
| Missing/Other | <1.0% | 1.3% |  | 9 | 6.9% |  | **29** |
| **Race %** |  |  | <0.0001 |  |  | <0.0001 |  |
| Black /African American | 59.7% | 54% |  | **11** | 28.2% |  | **54** |
| White | 35.7% | 36.6% |  | 2 | 51.4% |  | **30** |
| Asian | 0.1% | 1.0% |  | **12** | 1.2% |  | 2 |
| Native Hawaiian/Pac. Islander | <1.0% | <1.0% |  | 1 | <1.0% |  | 2 |
| American Indian/Alaska Native | <1.0% | <1.0% |  | 3 | <1.0% |  | 1 |
| Unknown | 3.6% | 7.2% |  | **16** | 18.2% |  | **34** |
| **Ethnicity %** |  |  | <0.0001 |  |  | <0.0001 |  |
| Not Hispanic or Latino | 96.5% | 92.3% |  | **19** | 80.6% |  | **35** |
| Hispanic or Latino | 1.5% | 2.9% |  | 9 | 2.9% |  | <1 |
| Unknown | 1.9% | 4.9% |  | **16** | 16.5% |  | **38** |

Table 4. Demographic Data

Table 4 displays several differences among the NLP Only and All ICD groups. Patients in the All ICD group were less likely to be married, and more likely to be divorced or separated, than the NLP Only group. Patients in the All ICD group were more likely to be male and younger. Patients in the NLP Only group were more likely to be female. Other significant variable attributes, according to ASD, include race (Black/African American, Asian, Unknown) and ethnicity (Not Hispanic or Latino, Unknown).

|  | All ICD | NLP Only | p-value (NLP Only vs All ICD) | ASD (%)(NLP Only vs All ICD) | No Problematic Opioid Use (%) | p-value (NLP Only vs No Problematic Opioid Use) | ASD (NLP Only vs No Problematic Opioid Use) |
|---|---|---|---|---|---|---|---|
| N | 6997 | 57331 |  |  | 158043 |  |  |
| Comorbidities (on or after patient entered cohort) |  |  |  |  |  |  |  |
| Hypertension | 57.1% | 53.5% | <0.0001 | 7 | 45.8% | <0.0001 | 15 |
| Diabetes Mellitus | 22.1% | 25.0% | <0.0001 | 7 | 20.4% | <0.0001 | 11 |
| Depression | 61.8% | 41.1% | <0.0001 | 42 | 19.6% | <0.0001 | 48 |
| Post-Traumatic Stress Disorder | 39.6% | 25.1% | <0.0001 | 31 | 10.8% | <0.0001 | 38 |
| Cancer | 9.0% | 12.1% | <0.0001 | 10 | 12.9% | <0.0001 | 2 |
| Tobacco | 62.0% | 31.1% | <0.0001 | 65 | 15.4% | <0.0001 | 38 |
| Alcohol | 60.9% | 23.8% | <0.0001 | 81 | 9.2% | <0.0001 | 44 |
| Other Drug Addictions | 66.6% | 18.6% | <0.0001 | 111 | 4.2% | <0.0001 | 47 |
| Traumatic Brain Injury | 11.5% | 7.0% | <0.0001 | 16 | 3.6% | <0.0001 | 15 |
| Anxiety | 39.7% | 27.6% | <0.0001 | 26 | 14.1% | <0.0001 | 34 |
| Neck Pain | 39.4% | 31.3% | <0.0001 | 17 | 18.2% | <0.0001 | 31 |
| Back Pain | 57.0% | 47.2% | <0.0001 | 20 | 30.7% | <0.0001 | 34 |
| Prior VA opioid prescription | 71.5% | 51.6% | <0.0001 | 42 | 32.1% | <0.0001 | 40 |
| Concurrent Benzodiazepine prescriptions | 19.4% | 10.1% | <0.0001 | 26 | 3.9% | <0.0001 | 24 |
| Mean/Standard Deviation Outpatient Encounters (since patient entered cohort; max 1 per day) | 50.9/49.8 | 33.4/31.3 | <0.0001 | 42 | 16.2/23.2 |  | 62 |

Table 5. Comorbidity, opioid prescription history, concurrent benzodiazepine prescription data, and means, standard deviations of outpatient encounters

Comorbidity data (Table 5) reveal notable trends, where, with the exception of cancer, hypertension, and diabetes, the All ICD group has significantly higher comorbidity levels compared to the NLP Only group according to ASD. This same trend is evident for prior opioid prescriptions, concurrent benzodiazepine prescriptions, and outpatient visits during the study period. All values in Table 5 reveal a continuum of escalating comorbidity levels, prior opioid and current benzodiazepine prescriptions, and care encounters according to group type.

*NLP Classifications among Notes*

Table 6 includes patient counts on positive snippet classifications and mean positive notes for the NLP Only patients, and the patients identified through ICD code that also had clinical notes positive for problematic opioid use (NLP/ICD, n = 6,243). For the NLP Only group, a positive classification was more likely to be incurred by a note containing a key phrase like 'opioid abuse' or 'withdrawal', than a note containing a specific drug name like 'hydrocodone'. However, the NLP/ICD group document counts in these categories were more similar. The mean positive snippet count for the NLP/ICD group is almost twice that of the NLP Only group.

| NLP Only, NLP/ICD Patient Groups, Positive Snippet Classifications | | |
|---|---|---|
| | Count of patients having positive snippets with Specific Drug Name | Count of patients having positive snippets with Other Key Phrase |
| NLP Only | 15,495 | 54,856 |
| NLP/ICD | 5,298 | 6,175 |
| | **Positive Snippets Mean/Standard Deviation** | |
| NLP Only | 1.7/1.5 | |
| NLP/ICD | 3.0/2.9 | |

Table 6. Patient counts by note key phrase type for patients with NLP positive notes; positive note count mean and standard deviation for patients with NLP positive notes

*Predominant Note Types*

There was also variation in the frequent positive clinical note types for the NLP Only patients, and the patients identified through ICD code that also had clinical notes positive for problematic opioid use (NLP/ICD). Figures 3 and 4 include bar charts of the 25 most frequent note types for each group.

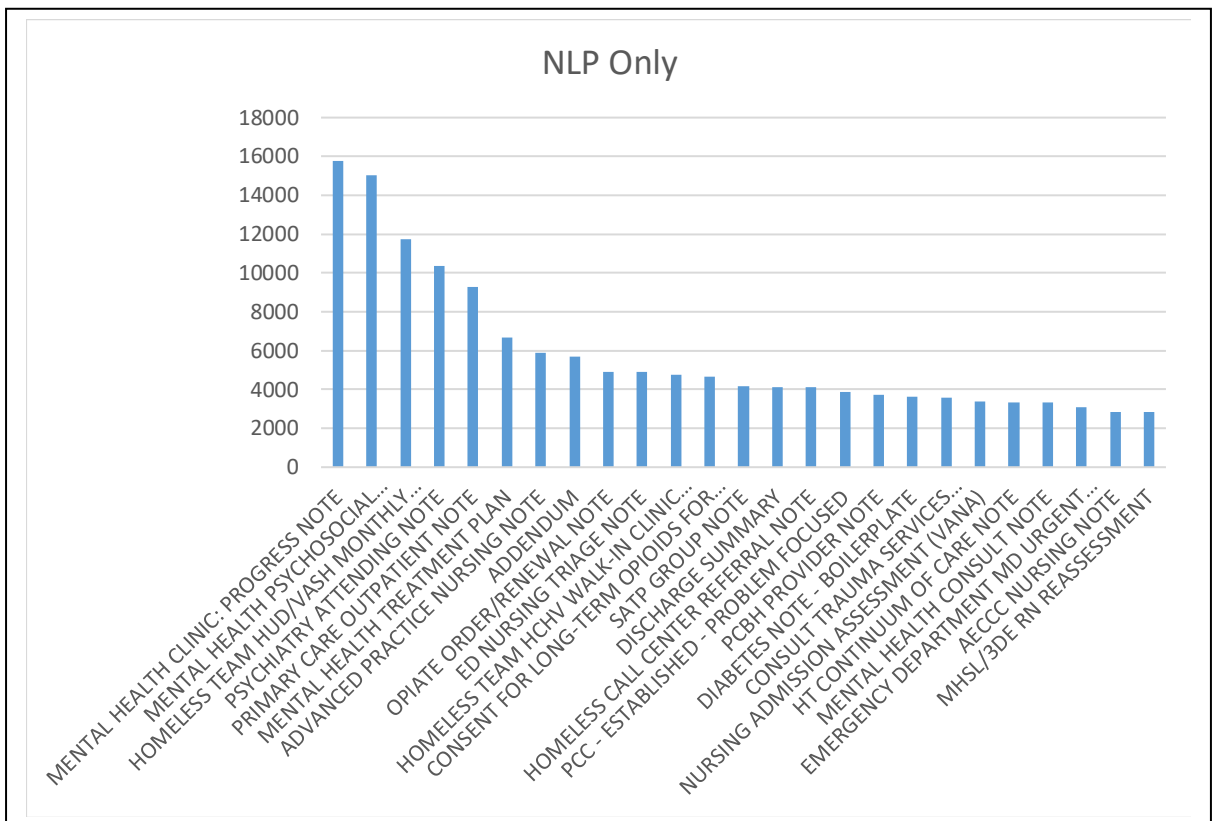

Figure 3. Prominent positive note types for NLP Only group

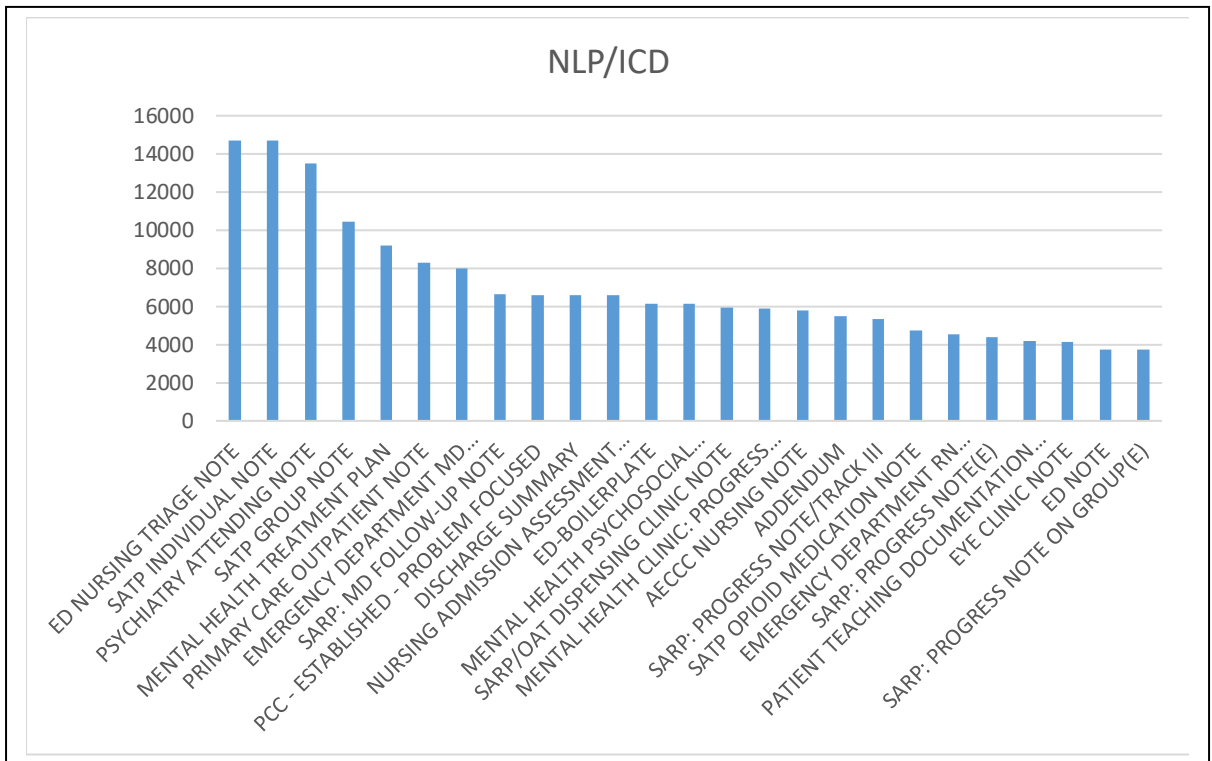

Figure 4. Prominent positive note types for patients receiving an ICD code who also had positive notes

The most frequent documents for the NLP Only group (Figure 3) address unique issues like opioid prescribing, and homelessness. Both charts include note types for substance abuse treatment (SATP i.e., substance abuse treatment program; SARP i.e., substance abuse rehabilitation program), but there are several in the NLP/ICD group (Figure 4). Both charts include several mental health-oriented notes, emergency department notes, and notes for other health issues (diabetes in the NLP Only group; eye clinic in the NLP/ICD group).

**Discussion**

This research produced an effective method to identify opioid risks among all patients. Our findings confirmed there were differences between patients receiving an OUD ICD code (All ICD group) and those having problematic opioid use documented only in clinical notes (NLP Only group). By patient counts, these findings also suggest that problematic opioid use is more likely to be documented in clinical notes than through ICD diagnostic codes.

The results indicate differences between the NLP Only and All ICD patient groups in regard to demographics (Table 4). Veterans in the All ICD group were significantly younger, and both groups were younger than the mean age of 58.3 of patients having no positive NLP or ICD classification. This finding correlates with previous research [19, 20]. Veterans within the NLP Only group were more likely to be female. Other research has found gender differences in opioid misuse communications, with women more likely than men to report opioid issues [21]. This difference may also be reflected in what providers document in clinical notes. Women are also more likely to receive opioid prescriptions on outpatient visits than men [22], and are more likely to be prescribed opioids for chronic conditions [23]. The patients in the All ICD group were less likely to be married. Chronic pain sufferers who are not married are more likely to abuse opioids [24], possibly due to less social support. The issue of race in opioid misuse and abuse must be examined through a contextual lens. Issues such as bias [25], and inequities of treatment and resources [26-28] must be considered in this larger societal discussion. In this study, race was a significant factor; this should also be interpreted within the framework of potential bias and access inequities. All of the significant demographic findings warrant future study.

There are notable differences in comorbidities between the All ICD and NLP Only groups (Table 5). This analysis includes a variety of comorbidities, including pain-oriented, mental health issues, addictions of other substances, and chronic diseases. The correlation between pain and opioid use issues had been studied [24], as well as between opioid abuse and misuse of other substances [29]. This current study highlights the differences between prevalence of these comorbidities in terms of the methods clinicians use to document problematic opioid use. Patients in the All ICD group were also more likely to have concurrent benzodiazepine prescriptions, prior opioid prescriptions, and receive more overall outpatient treatment. Levels in the All ICD and NLP Only groups exceed those of the No Problematic Opioid Use group. Collectively, the data suggest that the NLP Only group is worse off than patients having no documented problematic opioid use, and is approaching the state of the All ICD group. This is especially concerning, since the NLP Only patients have no recorded ICD code to enable providers in recognizing their opioid-related issues.

The findings in Table 6 and Figures 3 and 4 suggest different clinical note documentation patterns. For patients without a recorded OUD ICD diagnostic code, providers tended to discuss problematic opioid use outside of the context of specific drugs. For those with an OUD ICD diagnostic code, providers documented problematic opioid use in the context of specific drug names almost as often as by using other types of key phrases. Providers were also more likely to document problematic opioid use in clinical notes once an OUD ICD code was recorded. It is possible that providers felt more comfortable recording problematic opioid use once a structured data element was on record. Note types varied also by group. In the NLP Only group, notes addressing opioid prescribing and homelessness were prominent, whereas notes from substance abuse treatment groups were more prominent in the NLP/ICD group. This last observation suggests that patients receiving an ICD code are more likely to receive treatment.

These findings also suggest subgroups among Veterans for increased awareness of potential opioid misuse concern. These subgroups include women, people who are not married, and people of color. As a part of this awareness, there should also be cognizance of potential bias and inequities of available OUD resources for the patient, and a plan to address this.

Finally, our findings suggest that clinicians may be reluctant to code for opioid use disorder due to perceived negative implications by both clinicians and patients, as evidenced by the patient counts in the NLP Only and All ICD groups. The study took place in the greater Washington DC area where there are a large percentage of Veterans and civilians having formal associations with government, including employment. There may be a perception that having a diagnosis of OUD in the medical record may jeopardize these relationships. It is therefore incumbent on the healthcare team to search for documentation of opioid within the clinical notes.

*Limitations*

The NLP tool performance was tested on a set of VA clinical notes. The generalizability to other notes from other clinical settings will be tested in the next step of our research. Overall, VA patients are skewed in age, gender, and possibly race and other characteristics. This also may affect the generalizability of the findings; however, the younger Veteran population (opioid users and abusers tend to be younger [30]) is more like that of the general U.S. population.

*Future Work*

This work is part of a larger project addressing concordant care [31]. As part of a larger project, we will test the NLP tool developed using VA data on MedStar Health System data. MedStar Health is a large healthcare system in the Mid-Atlantic region, with 491 facilities, including 10 hospitals, and many clinics. We plan to further refine the NLP tool as needed. The NLP tool from this study has been made available as open-source software. Additional revisions will be posted in the future.

*Funding*

This work is funded by United States Veteran Affairs HSRD IIR grant HX003100-01A2.

*Ethical Approval*

No patients were contacted, only data from the EHR database were used.

## Conclusion

We developed an efficient NLP tool to detect problematic opioid use and applied it to clinical notes in order to better understand differences between patients with relevant unstructured documentation, as compared to those with a structured OUD diagnosis, in EHR records. We found significant differences in terms of demographic and comorbidity characteristics, prescription, outpatient visits, and documentation patterns. The findings suggest differences by medium – i.e., structured data or unstructured clinical notes – in how clinicians document problematic opioid use in patients.

## Conflict of Interest Statement

The authors have no conflict of interest to disclose.

## Data Availability Statement

The data used in this study are not publicly available because they include protected health information.

# Appendix

| Keywords/Key Phrases Occurrence Counts in Notes | | | | | |
|---|---|---|---|---|---|
| abstral: 4 | duragesic: 2,950 | hysingla: 75 | methadose: 46 | oxaydo: 6 | withdrawal: 975,669 |
| actiq: 587 | exalgo: 92 | kadian: 173 | morphine: 478,508 | oxycodone: 1,359,993 | zohydro: 79 |
| demerol: 25,370 | fentanyl: 221,602 | lorcet: 145 | norco: 1,555 | oxycontin: 72,111 | opioid dependence: 241,746 |
| dependence: 2,163,120 | fentora: 126 | lortab: 2,888 | opiate: 402,730 | percocet: 581,122 | polysubstance abuse: 102,912 |
| dilaudid: 115,808 | hydrocodone: 199,524 | meperidine: 2,788 | opiate abuse: 42,273 | roxicet: 298 | substance abuse: 1,403,341 |
| dolophine: 15 | hydromorphone: 212,451 | methadone: 648,154 | opioid: 899,720 | vicodin: 66,044 | substance dependence: 38,911 |

Table A. Key phrase occurrences in snippets; some snippets contained multiple key phrases.

| Keywords/Key Phrases Occurrence Counts by Patient | | | | | |
|---|---|---|---|---|---|
| abstral: 2 | duragesic: 1,241 | hysingla: 9 | methadose: 15 | oxaydo: 3 | withdrawal: 69,210 |
| actiq: 4 | exalgo: 29 | kadian: 33 | morphine: 25,252 | oxycodone: 41,943 | zohydro: 12 |
| demerol: 5,952 | fentanyl: 36,664 | lorcet: 24 | norco: 851 | oxycontin: 10,197 | opioid dependence: 4,482 |
| dependence: 54,712 | fentora: 10 | lortab: 1,201 | opiate: 26,358 | percocet: 37,951 | polysubstance abuse: 8,031 |
| dilaudid: 16,515 | hydrocodone: 20,937 | meperidine: 352 | opiate abuse: 2,500 | roxicet: 96 | substance abuse: 77,072 |
| dolophine: 11 | hydromorphone: 11,675 | methadone: 20,805 | opioid: 27,979 | vicodin: 18,326 | substance dependence: 11,327 |

Table B. Key phrase occurrences by patients, in the snippets.